\pdfoutput=1
\documentclass{bmvc2k}

\usepackage{booktabs}
\usepackage{xcolor}
\usepackage{float}

\title{Detecting Common Golfing Faults Using AI and Video-Derived Biomechanical Features}

\addauthor{Oliver Martin}{}{1}
\addauthor{Jeff Clark}{}{1}
\addauthor{Nawid Keshtmand}{nawid.keshtmand@bristol.ac.uk}{1}

\addinstitution{
School of Engineering Mathematics and Technology\\
University of Bristol\\
Bristol, UK
}

\runninghead{Martin, Clark, Keshtmand}{Detecting Common Golfing Faults Using AI}

\begin{document}
\maketitle

\begin{abstract}

Understanding how an athlete moves can improve both technique and safety, yet outside the laboratory technique is still assessed by eye. We present a pipeline that detects named golf swing faults from ordinary video, without markers or calibration. Instead of learning end to end, it tracks body pose and the clubhead in a body-centred frame, and classifies the biomechanical quantities a coach would measure, such as the height of the clubhead above the swing plane, with shallow boosted trees. We demonstrate the approach on two common faults, coming over-the-top (OTT) and early-extension (EE). On 160 swings and 143 swings from public sources it detects OTT and EE with a cross-validated ROC-AUC of 0.975 and 0.911 respectively. Because every feature is a biomechanical quantity a coach already uses, measured at a known phase of the downswing, each decision can be traced to the signals and the moments of the swing that drove it. The signals the model relies on most are the same ones a coach checks when diagnosing these faults. The pipeline processes phone video on a consumer laptop in seconds, so it could feasibly be developed into an app for golfers of all abilities and extended to other sports in which technique is judged against a biomechanical reference.

\end{abstract}

\section{Introduction}
\label{sec:intro}

How an athlete moves determines how well they perform, how likely they are to be injured, and in some sports whether their technique is legal at all. In golf, swing kinematics can differ between professionals and amateurs~\cite{zheng2008proam}, and the mechanics of the swing are implicated in low back pain, among the most common injuries in the sport~\cite{mchardy2006injuries,lindsay2014lowback,cole2016lowback}. Most of what is known about the technique comes from marker-based motion capture in the laboratory~\cite{bourgain2022review}, which is precise but requires a capture volume, markers, and an operator, and may not be representative of typical outdoor golfing conditions. Outside the laboratory, coaches assess technique by eye against a small set of key swing parameters~\cite{smith2015coaches}, and golfers without a coach are left to assess their own technique by filming themselves on a phone. A tool that serves professionals and amateurs alike must therefore work from a single ordinary video, with no markers, calibration, or special equipment.

\begin{figure}[!htbp]
\centering
\includegraphics[width=0.75\linewidth]{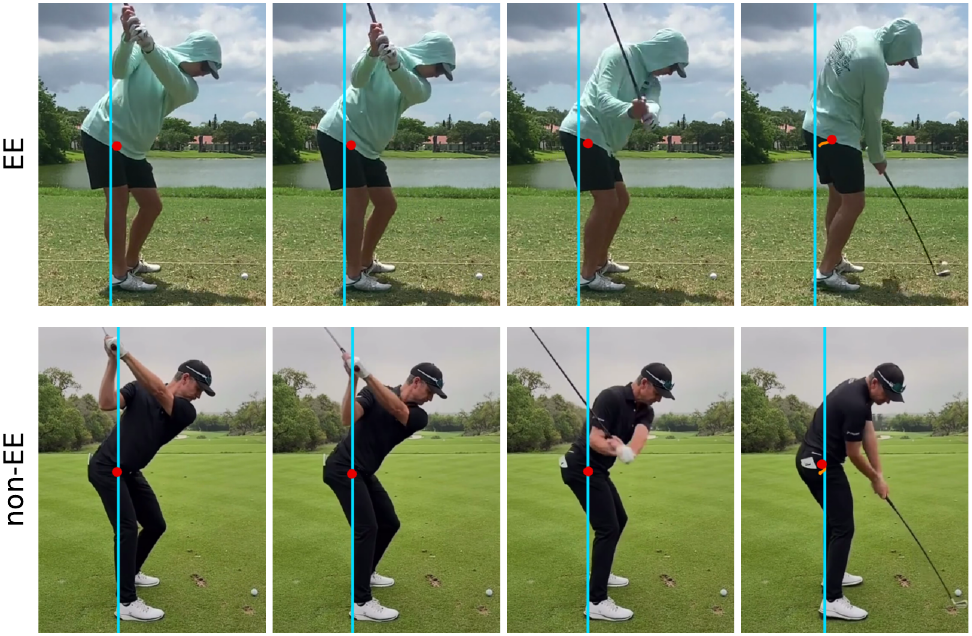}
\caption{Progression of an early-extension (EE) swing (top) and a non-EE swing (bottom), shown at the top of the backswing, one-third and two-thirds of the way through the downswing, and impact. Orange: tracked hip path (red dot: start
position). Cyan: the start line from the back of the foot. The hips of the amateur (top) start slightly right and extend far further than those of the experienced player (bottom).}
\label{fig:teaser_ee}
\end{figure}

Computer vision and machine learning are increasingly used in sports analytics, from player and ball tracking to motion analysis~\cite{thomas2017cvsports, turner2026markerless}, and markerless pose estimators~\cite{bazarevsky2020blazepose} now return body landmarks from phone video in real time. In golf, video has already been used for swing sequencing, self-training, and skill assessment~\cite{mcnally2019golfdb,liao2022aigolf,ingwersen2023handicap}, yet academic studies and consumer apps typically do not explicitly detect faults. In this paper we develop a pipeline to detect two named faults in golf technique:
\begin{enumerate}
    \item \textbf{Coming ``over-the-top'' (OTT)}: the club is thrown outward at the start of the downswing, so the clubhead travels above the swing plane. Perhaps the most common fault among high-handicap golfers, it typically produces a pull or a slice~\cite{tpi_ott}. Coaches diagnose it from the down-the-line view; so do we, from body pose and a tracked clubhead (Figure~\ref{fig:video_data}). It is a demanding test case: the decisive evidence is the club path, invisible to a pose estimator, judged over time against a biomechanical construct, the swing plane~\cite{coleman2005plane,kwon2012plane}.
    \item \textbf{Early-extension (EE)}: the hips thrust toward the ball during the downswing, so the torso stands up out of its address posture (Figure~\ref{fig:teaser_ee}). It typically produces a two-way miss, a block or a hook~\cite{tpi_ee}. Coaches diagnose it from the same view, by whether the hips keep their distance from the ball; so do we. 

\end{enumerate}

Our contributions are:
\begin{itemize}
\item A golf fault detection pipeline for ordinary monocular video, combining MediaPipe pose estimation with a YOLO11 clubhead detector fine-tuned on 700 hand-labelled frames and a gated Kalman/RTS tracker that bridges detection dropouts, utilising directly interpretable biomechanical features (Sec.~\ref{sec:method}).
\item A biomechanics-guided, phase-normalised \emph{trajectory} representation of the downswing, demonstrated on 160 and 143 swing videos, on which it detects OTT with a cross-validated ROC-AUC of 0.975 (Table~\ref{tab:main}) and EE with 0.911 (Table~\ref{tab:main_ee}).
\item Error and feature importance analyses for both faults; for OTT the single most informative signal is clubhead height above the swing plane (Table~\ref{tab:signals}).
\end{itemize}

\section{Related work}
\label{sec:related}

\paragraph{Technique assessment from pose estimation.}
Action quality assessment regresses judges' scores from whole-clip features, first from pose~\cite{pirsiavash2014aqa} and later from spatio-temporal RGB networks~\cite{parmar2019and}; it yields a global score rather than a decision about a named fault, and its benchmarks are broadcast footage of judged sports such as diving. Pose-based coaching systems compare a user's keypoints with exemplars and return corrective feedback~\cite{wang2019ai}. Closer to our task, named execution faults have been detected from monocular video with keypoint time series and compact classifiers: squat form classes from distance matrices~\cite{ogata2019temporal}, aberrant executions of the military press from time-series classifiers~\cite{singh2023fast}, and race-walking faults using pose estimation and machine learning models trained based on the judgements of multiple qualified judges~\cite{suzuki2024automatic}. These studies share our regime of small, expert-labelled datasets and interpretable features, but in each of them the fault is visible in the body alone, whereas OTT is a fault of the club path. Skeleton-based recognisers~\cite{yan2018spatial} learn from keypoint sequences end to end but need far more than 160 clips, and pose inputs have the further advantage of transferring to out-of-context actions where RGB models fail~\cite{weinzaepfel2021mimetics}.
\paragraph{Golf swing analysis from video.}
Monocular golf analysis has a long history in computer vision research. In 2003 \citet{gehrig2003golfclub} recovered the clubhead trajectory from ordinary swing video by fitting curves to club detections, and in 2005 \citet{urtasun2005golf} tracked the golfer's 3D pose through a swing with a learned motion model. Modern pipelines start from a learned body-pose estimator. GolfDB~\cite{mcnally2019golfdb} provides 1,400 professional swings annotated with eight event frames and the SwingNet event detector; we draw our negative class cases from it, and our top and impact detection plays the same role as its event detection, but from pose and clubhead tracks rather than RGB frames. Pose-based coaching tools synchronise a learner's swing with an expert's and visualise where the two diverge~\cite{liao2022aigolf}, estimate a golfer's handicap from video~\cite{ingwersen2023handicap}, or separate proficiency-related differences from individual style using 2D pose and club detection in a single camera view~\cite{yamamoto2023proficiency}. 3D pose estimation from monocular video has been found too inaccurate for golf kinematics~\cite{ingwersen2023monocular}. Closest in task to ours, \citet{feng2025scalstm} classify common swing errors from MediaPipe landmarks with an attention-augmented LSTM, on 160 swings recorded under controlled conditions, and \citet{kim2020poseedge} derive posture and tempo metrics and a projected club path from down-the-line footage, without a labelled fault to detect. Because body-pose estimators do not see the club, recent work fuses club and body keypoints to predict ball flight~\cite{jung2026graph}; we also treat the club as a first-class signal, with a dedicated detector and tracker, but use it to measure a named fault against a swing-plane line. To our knowledge no prior work detects OTT and EE automatically from monocular video.

\section{Method}
\label{sec:method}
Our representation is guided by interpretable biomechanical signals rather than learned end to end: we track the quantities a coach would measure and let a shallow classifier use them. The downswing lasts a fraction of a second, and both faults are defined by sequential movements, so features that summarise the whole swing discard most of the evidence. The pipeline therefore runs per-frame body pose and clubhead detection (Sec.~\ref{sec:pose}), maps the coordinates into a body-centred, aspect-corrected frame (Sec.~\ref{sec:norm}), detects the top of the backswing and impact (Sec.~\ref{sec:phases}), and resamples the downswing between them to a fixed number of phase steps, as gait cycles are time-normalised in biomechanics~\cite{helwig2011gait}. Within that window it tracks a small set of signals a coach would look at, including the height of the clubhead above a swing-plane line (Sec.~\ref{sec:traj}), and classifies their trajectories with boosted trees whose feature importance can be read off by signal and by phase (Sec.~\ref{sec:model}). EE detection reuses every stage except the clubhead detector.

\begin{figure}[!htbp]
\centering
\includegraphics[width=0.75\linewidth]{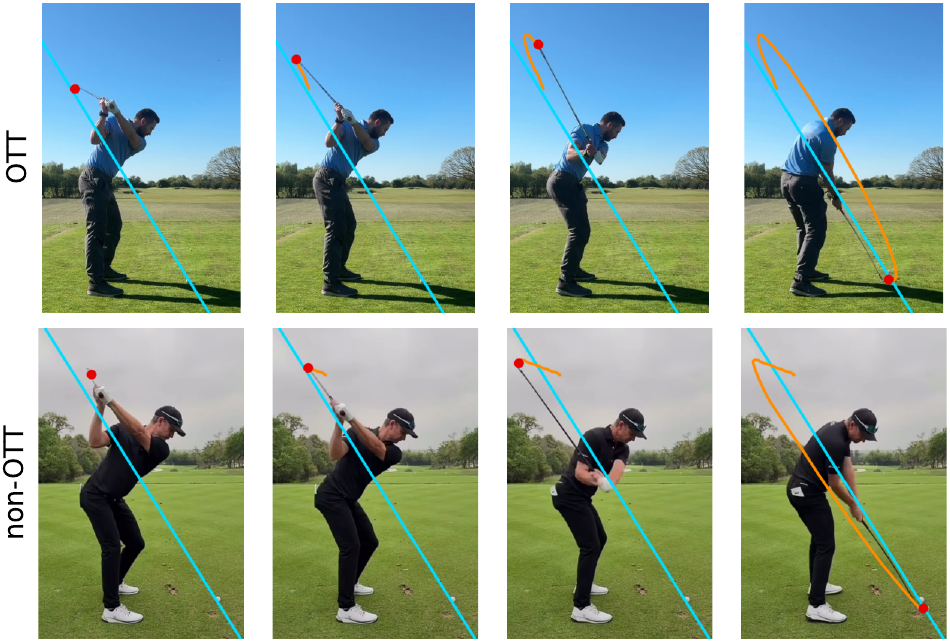}
\caption{Progression of an over-the-top (OTT) swing (top) and a non-OTT swing (bottom), shown at the top of the backswing, one-third and two-thirds of the way through the downswing, and impact. Orange: tracked clubhead path so far (red dot: current
position). Cyan: the swing-plane line from the clubhead at address to the trail elbow. The OTT clubhead loops outward and stays above the line, while the non-OTT clubhead drops onto the line before impact.}
\label{fig:video_data}
\end{figure}

\subsection{Body pose and clubhead detection}
\label{sec:pose}
Body landmarks are generated with MediaPipe Pose (33 landmarks per frame, video mode, model complexity~1) and are used by both fault models. The clubhead comes from a YOLO11-L detector (25.3M parameters) fine-tuned for three classes (ball, handle, clubhead) on 700 hand-labelled frames (560 train, 140 validation) for 80 epochs at 1280\,px. Inference runs at 640\,px with a confidence threshold of 0.10, followed by a recovery pass that re-detects missed clubheads in a zoomed crop around the predicted position; this raises the median fraction of frames with a detected clubhead from $\sim$62\% to $\sim$86\%. The raw clubhead track is then cleaned and smoothed. Detections more than 6.0 robust deviations (MAD) from the local median path, or more than 1.2 body heights from the lead wrist, are rejected. A constant-acceleration Kalman filter with confidence-scaled measurement noise (measurement and process noise variances $4{\times}10^{-4}$ and $5{\times}10^{-3}$) and a 3.5$\sigma$ innovation gate is followed by a Rauch--Tung--Striebel (RTS) smoother, which bridges gaps of up to 15 frames. These settings are fixed across all OTT experiments. Figure~\ref{fig:video_data} shows an OTT and a non-OTT swing with the tracked clubhead path and the swing-plane line. The clubhead detector and tracker serve the OTT model only: EE is a fault of the body rather than the club, so its model works from the body landmarks alone.

\subsection{Body-frame normalisation}
\label{sec:norm}
Pose and detector coordinates arrive as fractions of frame width and
height, which are different units on any non-square frame. Let $(x, y)$
be a keypoint's horizontal and vertical image position. We first multiply
$x$ by the frame aspect ratio $W/H$ so that both axes are in frame
heights, which removes the angle distortion of portrait frames. Each
keypoint is then mapped to
\begin{equation}
  \tilde{x} = \frac{s\,(x - x_{\mathrm{hip}})}{B}, \qquad
  \tilde{y} = \frac{y_{\mathrm{hip}} - y}{B},
\end{equation}
where $(x_{\mathrm{hip}}, y_{\mathrm{hip}})$ is the mid-hip, $B$ is body
height (the median shoulder-to-ankle distance over the clip), and
$s = \pm 1$ orients the horizontal axis by the backswing direction. This
decouples frame resolution, camera distance and the golfer's size, and
makes the signals invariant to handedness and camera side.

\subsection{Swing phases}
\label{sec:phases}
Both faults are faults of the downswing, so both models work on the window from the top of the backswing to impact, and both events must be located in every clip. Impact is the frame at which the clubhead returns to its address height (the ball) and turns upward; anchoring impact to the ball prevents a high finish, which comes after impact, from being mistaken for the top. The top is defined as the lead wrist's last up-to-down reversal before impact. It is anchored to the wrist rather than the clubhead because the clubhead's highest point is itself shifted by the OTT move: with a clubhead-defined top, the top fell at a different phase in each class. When the clubhead is tracked on fewer than half the frames, both events default to the wrist trajectory, and the EE model, which does not track the clubhead, locates both events from the wrist throughout.

\subsection{Classifier and features}
\label{sec:model}
Biomechanical signals are extracted from each clip, as positions and as their frame-to-frame differences, and classified with XGBoost.

\subsubsection{Absolute position and change in position}
\label{sec:traj}
The downswing is resampled to $N{=}30$ phase steps, and at each step we record a set of signals a coach would look at, in the body frame (Sec.~\ref{sec:norm}); their frame-to-frame differences $\Delta s_t = s_t - s_{t-1}$ double the feature count. For OTT the eleven signals are lead wrist $x,y$; trail elbow $x,y$; horizontal shoulder and hip separation; shoulder tilt; clubhead $x,y$; shaft angle; and the signed height of the clubhead above the swing-plane line (Sec.~\ref{sec:plane}), giving 330 position and 330 delta features. The last of these measures the defining sign of OTT directly, a clubhead that stays above the swing plane through the downswing~\cite{tpi_ott}, but not decisively. The other signals describe the hands, arms, club and torso around it and separate the classes at different moments.

For EE, a fault of the body rather than the club, the fifteen signals are the horizontal and vertical displacement of the hips and of the shoulders from address, measured relative to the mid-ankle; the head's rise from address; horizontal hip separation; the $x,y$ of both knees and both ankles; and the spine angle, the forward tilt of the mid-hip to mid-shoulder line, giving 450 position and 450 delta features. The hips' horizontal displacement measures the defining sign of EE directly, a pelvis that moves toward the ball during the downswing~\cite{tpi_ee}. The other signals describe what the hip thrust does to the torso, head and legs, and where the hips sit over the feet.

\subsubsection{Swing plane}
\label{sec:plane}
The plane line runs from the ball to the trail elbow, both taken at address in the first 0.5\,s of the clip, and is held fixed through the downswing (Figure~\ref{fig:video_data}). At address the clubhead rests on the ball, so its tracked position there gives the ball point, avoiding the ball detector, which on a driving range can jump between stray balls. Every swing begins the downswing well above this line, so its early frames carry little information.%

\subsubsection{Trajectories rather than summaries}
To assess the importance of temporal trajectories, we compare a model trained on the 30 separate phase steps with a baseline trained on scalar summaries of the whole downswing, taken from the coaching definition of each fault (Sec.~\ref{sec:exp}). The same eleven signals (fifteen for EE) enter the trajectory model as positions alone, as frame-to-frame deltas alone, and as both together (Sec.~\ref{section:OTT_results} and Sec.~\ref{sec:ee}).

\section{Experimental setup}
\label{sec:exp}

\paragraph{Data.}
\label{sec:data}
A dataset of 169 video clips was assembled for OTT detection. \emph{Positives}, which display the OTT fault, total 68 clips: 62 are amateur clips sourced from Reddit, 4 from YouTube and 2 from other sources. \emph{Negatives} (non-OTT, 101 clips) are 83 professional clips from GolfDB ($160{\times}160$\,px, cropped to the annotated swing plus 15 frames either side), 13 amateur clips from TikTok, 4 from Reddit and 1 from YouTube. All clips share the down-the-line camera angle and were checked for a consistent view and adequate lighting. Nine clips (5 OTT, 4 non-OTT) whose detected downswing window (Sec.~\ref{sec:phases}) does not actually descend (cut off or flat) are discarded, since their phase-normalised features would describe no real downswing, leaving 160 clips for all OTT experiments. %

For EE detection, 143 video clips were used, consisting of 60 amateur clips from the OTT dataset graded independently for EE, 28 extra Reddit clips posted about EE, and 55 GolfDB professional clips, assumed free of it. In total 46 clips were positive for EE. These 143 clips remain after excluding 56 clips whose hip movement is not tracked effectively.

\paragraph{Classifier.}
\label{sec:clf}
Both fault models use the position and delta trajectory features of Sec.~\ref{sec:traj} (660 for OTT, 900 for EE), classified with XGBoost: depth-2 trees, 400 estimators, learning rate 0.05, L2 regularisation $\lambda{=}3$, and column subsampling of $\min(1, 8/\sqrt{d})$ for $d$ features. Shallow trees keep the feature importance readable by signal and by phase. The same settings are used for OTT and EE.

\paragraph{Scalar baseline.} For comparison, baseline models for both faults were trained on a small number of scalar features from the coaching definition of each fault. For OTT the baseline scalar model utilises four features: the horizontal gap between the lead hand's downswing and backswing paths; how far the hand moves outward relative to how far it drops early in the downswing; the early change in shaft angle; and the clubhead's height above the swing plane, weighted toward mid-downswing (Sec.~\ref{sec:plane}). The EE scalar baseline uses three features: how far the hips move toward the ball from top to impact; how far they sit past the ankles at impact compared with address; and the loss of forward spine tilt from address to impact. For both, an XGBoost classifier was trained with 200 trees of depth 3 and learning rate 0.1.
\paragraph{Evaluation protocol.}
Repeated stratified 5-fold cross-validation (3 repeats, seed 42), reporting mean $\pm$ s.d.\ of ROC-AUC over the 15 folds, and F1, precision and recall at a fixed 0.5 threshold. All results in Sec.~\ref{sec:results} use this protocol unless stated otherwise; the `full model' for OTT or EE is the one using both the positions and the deltas of the trajectory.

\section{Results and discussion}
\label{sec:results}
\subsection{Over-the-top (OTT)}
\subsubsection{Classification of ``over-the-top'' (OTT)}
\label{section:OTT_results}
Table~\ref{tab:main} summarises detection performance on the 160 swings. The full trajectory model (positions and deltas, 660 features) achieves the best performance and reaches a cross-validated ROC-AUC of $0.975\pm0.019$. In practical terms, at the default 0.5 decision threshold, when the model flags a swing as OTT it is correct in $\sim$93\% of cases, and it detects $\sim$89\% of the OTT swings in the dataset. The four scalar features that summarise the whole downswing reach only $0.874\pm0.046$, so resolving the downswing in time accounts for most of the improvement. Positions and deltas each perform well alone (AUC of 0.962 and 0.969) and are complementary when combined.

\begin{table}[!htbp]
\centering
\small
\begin{tabular}{lrcccc}
\toprule
Features & Dim. & AUC & F1 & Prec. & Rec. \\
\midrule
Scalar baseline        & 4   & 0.874\,$\pm$\,0.046 & 0.760 & 0.813 & 0.740 \\
Trajectory: positions  & 330 & 0.962\,$\pm$\,0.031 & 0.884 & 0.931 & 0.847 \\
Trajectory: deltas     & 330 & 0.969\,$\pm$\,0.020 & 0.840 & 0.881 & 0.821 \\
Trajectory: pos+delta  & 660 & \textbf{0.975}\,$\pm$\,\textbf{0.019} & 0.908 & 0.931 & 0.894 \\
\bottomrule
\end{tabular}
\caption{OTT fault detection across 160 swings, 63 of them positive for OTT; ROC-AUC is mean\,$\pm$\,s.d.\ over 15 folds. F1, precision (``Prec.'') and recall (``Rec.'') are at the 0.5 decision threshold; ``Dim.'' is the number of features.}
\label{tab:main}
\end{table}

Figure~\ref{fig:curves} (Appendix~\ref{appendix:class_mean_trajectories}) shows the class-mean trajectory of each signal over the downswing. Clubhead height above the plane separates the classes most clearly, and the other club signals separate over most of the downswing. The body signals separate at their own moments: the lead wrist's horizontal position differs from the start, and shoulder tilt and hip separation differ slightly in the first half. The evidence is thus confined to particular phases, which the phase-resolved representation preserves and a whole-swing summary would average away.

\subsubsection{Feature importance by signal and phase}
\label{sec:importance}

\begin{figure}[!htb]
\centering
\subfigure[Full model]{%
  \includegraphics[width=\linewidth]{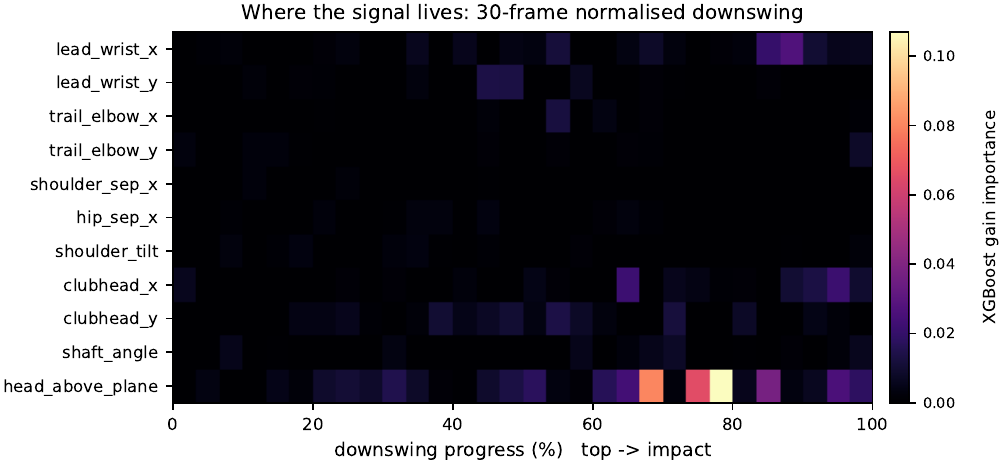}}\\
\subfigure[Body pose only]{%
  \includegraphics[width=\linewidth]{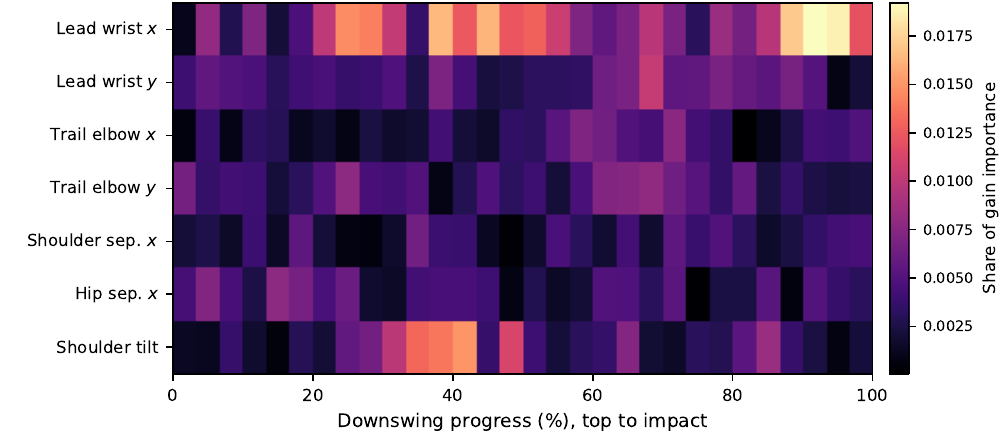}}
\caption{Feature importance of the OTT classifier by signal (rows) and downswing phase (columns; 0\% is the top of the backswing, 100\% impact), with each signal's position and delta features at a phase step combined into one cell. (a)~Full model. (b)~Body-pose-only model (AUC 0.883). Colour scales differ between the panels.} %
\label{fig:heatmaps}
\end{figure}

We examine which signals and which moments of the downswing the OTT model relies on in two ways. The first trains the classifier on a single signal, its 30 positions and 30 deltas, and reports the AUC that signal reaches alone (Table~\ref{tab:signals}), a measure of how informative each signal is regardless of the others. The second lays out a trained model's feature importance, the share of XGBoost's gain attributed to each feature, as a map of signal against downswing phase, for the full model and for a model trained on the body signals alone (Figure~\ref{fig:heatmaps}). %

\paragraph{Single-signal models.}
The ranking in Table~\ref{tab:signals} follows the definition of the fault. Clubhead height above the swing plane, the quantity a coach judges OTT by, is the strongest single signal with an AUC of 0.940, within 0.035 of the full model (Table~\ref{tab:main}), and the other three club signals follow: the clubhead's horizontal and vertical position (0.875 and 0.867) and the shaft angle (0.833). The body signals are weaker but far from uninformative. The lead wrist's horizontal position (0.820) outranks its height (0.689): in the down-the-line view the horizontal axis runs across the target line, the direction in which an OTT move throws the hands and club outward. Hip separation reaches 0.789 although it barely differs between the classes in any third of the downswing; the trail elbow (0.693 and 0.687) and shoulder tilt (0.650) add little alone, and the horizontal shoulder separation (0.571) is close to chance. The body alone thus carries a usable signal of a fault defined by the club, and the full model's 0.975 exceeds the strongest club and body signals, so the two are partly complementary.

\begin{table}[!htbp]
\centering
\small
\begin{tabular}{lc@{\hspace{2em}}lc}
\toprule
Signal & AUC & Signal & AUC \\
\midrule
Head above plane & 0.940\,$\pm$\,0.026 & Trail elbow $x$    & 0.693\,$\pm$\,0.067 \\
Clubhead $x$     & 0.875\,$\pm$\,0.057 & Lead wrist $y$     & 0.689\,$\pm$\,0.107 \\
Clubhead $y$     & 0.867\,$\pm$\,0.043 & Trail elbow $y$    & 0.687\,$\pm$\,0.088 \\
Shaft angle      & 0.833\,$\pm$\,0.073 & Shoulder tilt      & 0.650\,$\pm$\,0.083 \\
Lead wrist $x$   & 0.820\,$\pm$\,0.071 & Shoulder sep.\ $x$ & 0.571\,$\pm$\,0.063 \\
Hip sep.\ $x$    & 0.789\,$\pm$\,0.077 & & \\
\bottomrule
\end{tabular}
\caption{Per-signal cross-validated ROC-AUC (mean\,$\pm$\,s.d.\ over 15 folds, 160 swings) of the OTT classifier trained on each signal individually, i.e.\ its 30 phase-step positions and 30 deltas (60 features), under the
protocol of Sec.~\ref{sec:exp}. ``Head above plane'' is the signed height of the clubhead above the swing-plane line (Sec.~\ref{sec:plane}); $x$ and $y$ are horizontal and vertical body-frame coordinates (Sec.~\ref{sec:norm}) and ``sep.'' the horizontal separation of the shoulders or hips. Signals are sorted by AUC; the full model using all
eleven signals reaches 0.975 (Table~\ref{tab:main}).}
\label{tab:signals}
\end{table}

\paragraph{Importance maps.}
For the full model the importance map is sparse (Figure~\ref{fig:heatmaps}a): almost all of the importance sits in clubhead height above the swing plane between about 65\% and 80\% of the downswing, with smaller contributions from the same signal in the last 20\% and from the clubhead's and lead wrist's horizontal position near impact. Contributions in the first half are minimal: every swing starts the downswing well above the plane line, and the classes diverge in the middle third and are furthest apart in the last. The body rows, by contrast, are nearly empty, which appears to contradict Table~\ref{tab:signals}, where the same body signals reach AUCs between 0.57 and 0.82 on their own. The two views measure different things. A single-signal AUC measures how much a signal carries by itself, whereas importance measures how much the training loss fell when the trees split on a feature, and boosting chooses at each split the feature that reduces the loss most. Clubhead height above the plane separates the classes better than any body signal, so the trees split on it almost every time, and the body signals, which see the same outward move from the body's side and are correlated with it, are rarely needed once it has been used. Their low importance means the club signal carries their information more directly, not that they carry none: the table shows what each signal knows, and the map what the model uses.

Removing the club signals makes this visible. A body-only model trained on the seven body signals (AUC 0.883) spreads its importance across signals and across the downswing (Figure~\ref{fig:heatmaps}b): the lead wrist's horizontal position from about 20\% to 55\% of the downswing and again near impact, shoulder tilt between about 30\% and 50\%, hip separation at the very start, and the wrist and trail-elbow heights in the second half.
In both maps the evidence is confined to particular phases rather than spread evenly, which is what a whole-swing summary discards, and in both the signals and moments the model relies on are those a coach looks for.

\subsection{Early-extension (EE)}
\subsubsection{Classification of early-extension (EE)}
\label{sec:ee}
The EE experiments test whether the same recipe transfers to a fault of the body rather than the club. %
Table~\ref{tab:main_ee} presents detection of EE on the 143 clips. With position and delta features the model reaches a cross-validated ROC-AUC of $0.911\pm0.065$ and, at the 0.5 threshold, an F1 of 0.706 with precision 0.852 and recall 0.619: a swing flagged as EE is correct in $\sim$85\% of cases, but only $\sim$62\% of EE swings are detected. As with OTT, precision exceeds recall, but the gap is wider. The classes are more imbalanced (46 of 143 clips, against 63 of 160 for OTT), which leaves the fixed 0.5 threshold on the conservative side; the AUC shows that the ranking of swings is better than the thresholded figures suggest, and a lower threshold would trade precision for recall. The fold-to-fold spread of AUC is also more than three times that of OTT ($\pm$0.065 against $\pm$0.019), as expected with $\sim$nine positives per test fold. As for OTT, a trajectory beats scalar summaries of the same movement.

\begin{table}[H]
\centering
\small
\begin{tabular}{lrcccc}
\toprule
Features & Dim. & AUC & F1 & Prec. & Rec. \\
\midrule
Scalar baseline        & 3   & 0.726\,$\pm$\,0.079 & 0.536 & 0.619 & 0.491 \\
Trajectory: positions  & 450 & 0.824\,$\pm$\,0.061 & 0.539 & 0.672 & 0.465 \\
Trajectory: deltas     & 450 & 0.883\,$\pm$\,0.080 & 0.727 & 0.853 & 0.649 \\
Trajectory: pos+delta  & 900 & \textbf{0.911}\,$\pm$\,\textbf{0.065} & 0.706 & 0.852 & 0.619 \\
\bottomrule
\end{tabular}
\caption{EE fault detection across 143 swings, 46 of them positive for EE; ROC-AUC is mean\,$\pm$\,s.d.\ over 15 folds. Trajectory rows use 15 hip, leg and torso signals. The 143 swings exclude 56 clips whose hip movement was not tracked effectively (Sec.~\ref{sec:data}), so these results are an upper bound. F1, precision (``Prec.'') and recall (``Rec.'') are at the 0.5 decision threshold; ``Dim.'' is the number of features.}
\label{tab:main_ee}
\end{table}

\subsubsection{Feature importance by signal and phase}
\label{sec:EE_importance}

\begin{figure}[!htb]
\centering
\includegraphics[width=\linewidth]{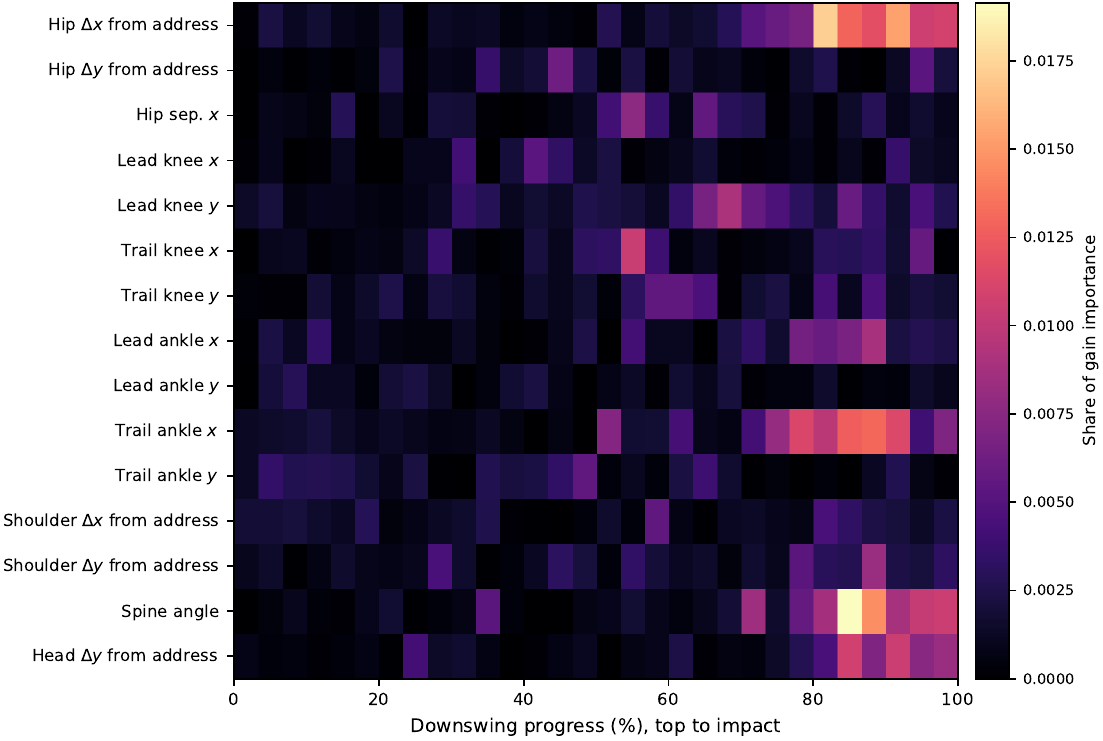}
\caption{Feature importance of the EE classifier of Table~\ref{tab:main_ee} (143 clips, AUC 0.911) by signal (rows) and downswing phase (columns; 0\% is the top of the backswing, 100\% impact), with each signal's position and delta features at a phase step combined into one cell. The model sees body landmarks only; the fifteen signals are defined in Table~\ref{tab:signals_ee}. Importance concentrates at 80--100\% of the downswing in the hips' horizontal displacement from address, the spine angle and the rise of the head.}
\label{fig:ee_heatmap}
\end{figure}

We examine which signals and moments of the downswing the EE model relies on in the same two ways as for OTT (Sec.~\ref{sec:importance}): the AUC each signal reaches alone (Table~\ref{tab:signals_ee}), and the trained model's share of gain by signal and downswing phase (Figure~\ref{fig:ee_heatmap}). As the EE model works from body landmarks alone, there is one map rather than two.

\paragraph{Single-signal models.}
The ranking in Table~\ref{tab:signals_ee} again follows the definition of the fault. The hips' horizontal displacement from address, the quantity a coach judges EE by (Sec.~\ref{sec:intro}), is the strongest signal at 0.910, as high as the full model's 0.911 (Table~\ref{tab:main_ee}), so the other fourteen signals add nothing measurable to the ranking of swings. Those that follow are consequences of the same movement: the spine angle (0.834) is the torso standing up out of its address tilt, and since the body frame is centred on the mid-hip (Sec.~\ref{sec:norm}), the trail and lead ankle $x$ (0.826 and 0.727) are the horizontal offset between hips and feet, that is, how far the hips have moved out over the ankles. The head's rise (0.687) and three knee coordinates (0.605 to 0.641) carry less, and the remaining seven signals are close to chance (0.516 to 0.584), among them the vertical displacement of the hips and shoulders, the shoulders' horizontal displacement and hip separation, which reached 0.789 for OTT. EE is therefore visible as the hips moving toward the ball under the shoulders and over the feet, and as the posture change that follows. 

\begin{table}[!htbp]
\centering
\small
\begin{tabular}{lc@{\hspace{2em}}lc}
\toprule
Signal & AUC & Signal & AUC \\
\midrule
Hip $\Delta x$ from address  & 0.910\,$\pm$\,0.057 & Hip $\Delta y$ from address      & 0.584\,$\pm$\,0.098 \\
Spine angle                  & 0.834\,$\pm$\,0.090 & Shoulder $\Delta x$ from address & 0.563\,$\pm$\,0.113 \\
Trail ankle $x$              & 0.826\,$\pm$\,0.045 & Lead ankle $y$                   & 0.559\,$\pm$\,0.083 \\
Lead ankle $x$               & 0.727\,$\pm$\,0.113 & Hip sep.\ $x$                    & 0.552\,$\pm$\,0.105 \\
Head $\Delta y$ from address & 0.687\,$\pm$\,0.102 & Shoulder $\Delta y$ from address & 0.531\,$\pm$\,0.073 \\
Trail knee $x$               & 0.641\,$\pm$\,0.079 & Lead knee $x$                    & 0.530\,$\pm$\,0.109 \\
Trail knee $y$               & 0.611\,$\pm$\,0.103 & Trail ankle $y$                  & 0.516\,$\pm$\,0.091 \\
Lead knee $y$                & 0.605\,$\pm$\,0.064 &                                  &                     \\
\bottomrule
\end{tabular}
\caption{Per-signal cross-validated ROC-AUC (mean\,$\pm$\,s.d.\ over 15 folds, 143 swings) of the EE classifier trained on each signal individually, i.e.\ its 30 phase-step positions and 30 deltas (60 features), under the
protocol of Sec.~\ref{sec:exp}. ``$\Delta x$'' and ``$\Delta y$'' are horizontal and vertical displacement from the address position, relative to the mid-ankle; ``Spine angle'' is the forward tilt of the mid-hip to mid-shoulder line; the other $x$ and $y$ are horizontal and vertical body-frame coordinates (Sec.~\ref{sec:norm}) and ``sep.'' the horizontal separation of the hips. Signals are sorted by AUC; the full model using all
fifteen signals reaches 0.911 (Table~\ref{tab:main_ee}).}
\label{tab:signals_ee}
\end{table}

\paragraph{Importance maps.} Figure~\ref{fig:ee_heatmap} shows what the EE model relies on. Through the first three quarters of the downswing importance is low and scattered, with only hip separation and the knees carrying weak, isolated importance in its middle. Importance concentrates at 80--100\% of the downswing in the hips' horizontal displacement from address, the spine angle and the rise of the head, with the horizontal position of the trail ankle alongside. This is the fault as a coach defines it (Sec.~\ref{sec:intro})~\cite{tpi_ee}: the hips move toward the ball and the torso stands up out of its address posture, and the evidence appears in the approach to impact rather than at the top. As for OTT, the decisive signals are confined to one part of the downswing, here its last fifth, which the phase-resolved representation keeps separate and a whole-swing summary would average away. The two views agree: the signals that carry the importance in the last fifth are four of the five strongest single signals in Table~\ref{tab:signals_ee}, and the rows that stay dark throughout are the signals near chance. That the gain is spread over the spine, ankle and head rows while the full model is no better than the hips' displacement alone follows the reading of gain in Sec.~\ref{sec:importance}: correlated views of one movement share it.

\subsection{Why OTT is detected better than EE}
EE is harder to detect than OTT for three reasons. First, its defining movement is small and overlaps between classes: at impact, the median EE swing has moved its hips 0.033 body heights toward the ball against 0.003 for non-EE amateurs, but a quarter of EE swings move less than a quarter of non-EE swings, whereas clubhead height above the plane differs by 0.32 body heights in the last third of the downswing. Second, the clubhead is tracked directly, while the hip is a joint centre inferred by the pose estimator. Third, EE is graded on a continuum and has fewer positives. %

\subsection{Runtime}
Run time was assessed on an Apple M2 laptop (Table~\ref{tab:runtime}). With the YOLO11-L detector used throughout, the forward pass takes 281\,ms/frame, and full clubhead tracking of a 104-frame clip, including video decoding and the recovery pass, takes 12.7\,s (42.3\,s on the CPU). A phone clip can therefore be scored locally on consumer hardware within seconds of the swing, while it is still fresh; because OTT and EE are properties of the whole downswing, feedback is per swing rather than per frame. The smaller YOLO11-S and YOLO11-N variants run at 38 and 56 frames per second, which would allow near-instantaneous feedback once their detection accuracy has been validated.

\begin{table}[!htbp]
\centering
\small
\begin{tabular}{llrrr}
\toprule
Detector & Device & ms/frame & fps & 3\,s clip (s) \\
\midrule
YOLO11-L$^\dagger$ & CPU       & 1374 & 0.7 & 123.7 \\
YOLO11-L$^\dagger$ & GPU (MPS) & 281  & 3.6 & 25.3 \\
\midrule
YOLO11-S           & GPU (MPS) & 26   & 38  & 2.3 \\
YOLO11-N           & GPU (MPS) & 18   & 56  & 1.6 \\
\bottomrule
\end{tabular}
\caption{Detector forward-pass time per frame on an Apple M2 laptop for a synthetic $640{\times}640$ input. ``fps'' is frames processed per second (1000/ms); typical video is 30\,fps. ``3\,s clip'' is the processing time for 90 frames. $^\dagger$The detector used in all experiments; the smaller S (``small'') and N (``nano'') variants were timed only and their detection accuracy is untested.}
\label{tab:runtime}
\end{table}

\section{Conclusion}
\label{sec:limits}
We have presented a pipeline that detects named golf swing faults from a single ordinary video, with no markers, calibration or special equipment, and applied it to OTT and EE. Rather than learning end to end, it tracks the biomechanical quantities a coach would measure, including the height of the clubhead above a swing-plane line, in a body-centred frame, and shallow boosted trees on these trajectories detect OTT and EE with a cross-validated ROC-AUC of 0.975 and 0.911 respectively. What the models rely on agrees with coaching practice: for OTT the most informative signals are the clubhead's height above the swing plane and its position, and the classes separate at distinct moments of the downswing that a whole-swing summary averages away, while for EE the movement of the hips is the driving factor. Because the signals are the ones a coach uses, a detection can be explained to the golfer in the same terms, which is what a coaching or injury-prevention tool needs. The study has several limitations. The dataset is small, only two faults are covered, and the clips come from a single down-the-line view in which hip rotation is foreshortened. Fault labels are partly confounded with skill level: most non-faulty clips are professional swings from GolfDB, while nearly all positives are amateurs, so the model may partly learn differences between professional and amateur swings rather than the fault itself. The static-frame probe (Appendix~\ref{app:video}) suggests it does not rely heavily on pre-fault cues, but does not rule the confound out. No qualified coach selected or graded the clips, so some labels are likely to be noisy, particularly for EE, which is graded on a continuum. Future work will expand the dataset from public sources, with more non-faulty amateur swings so the skill-level confound can be tested directly, have multiple qualified coaches grade the clips and report their inter-rater agreement, and assess further faults. Single-image 3D body reconstruction such as SAM 3D~\cite{chen2026sam} may recover the rotations that a 2D view hides. Clubhead tracking already runs in about 13\,s per clip on an Apple M2, so an app that returns fault feedback from a phone recording is within reach, and the same recipe of coach-defined signals, phase-normalised over a detected movement window, applies to other sports in which technique is judged against a biomechanical reference.

\paragraph{Acknowledgments}
OM was funded by the University of Bristol Faculty of Science and Engineering Summer Internship Scheme 2026.

\newpage
\bibliography{refs}

\newpage
\appendix

\section{Class-mean trajectories for OTT}
\label{appendix:class_mean_trajectories}

\begin{figure}[!htbp]
\centering
\includegraphics[width=\linewidth]{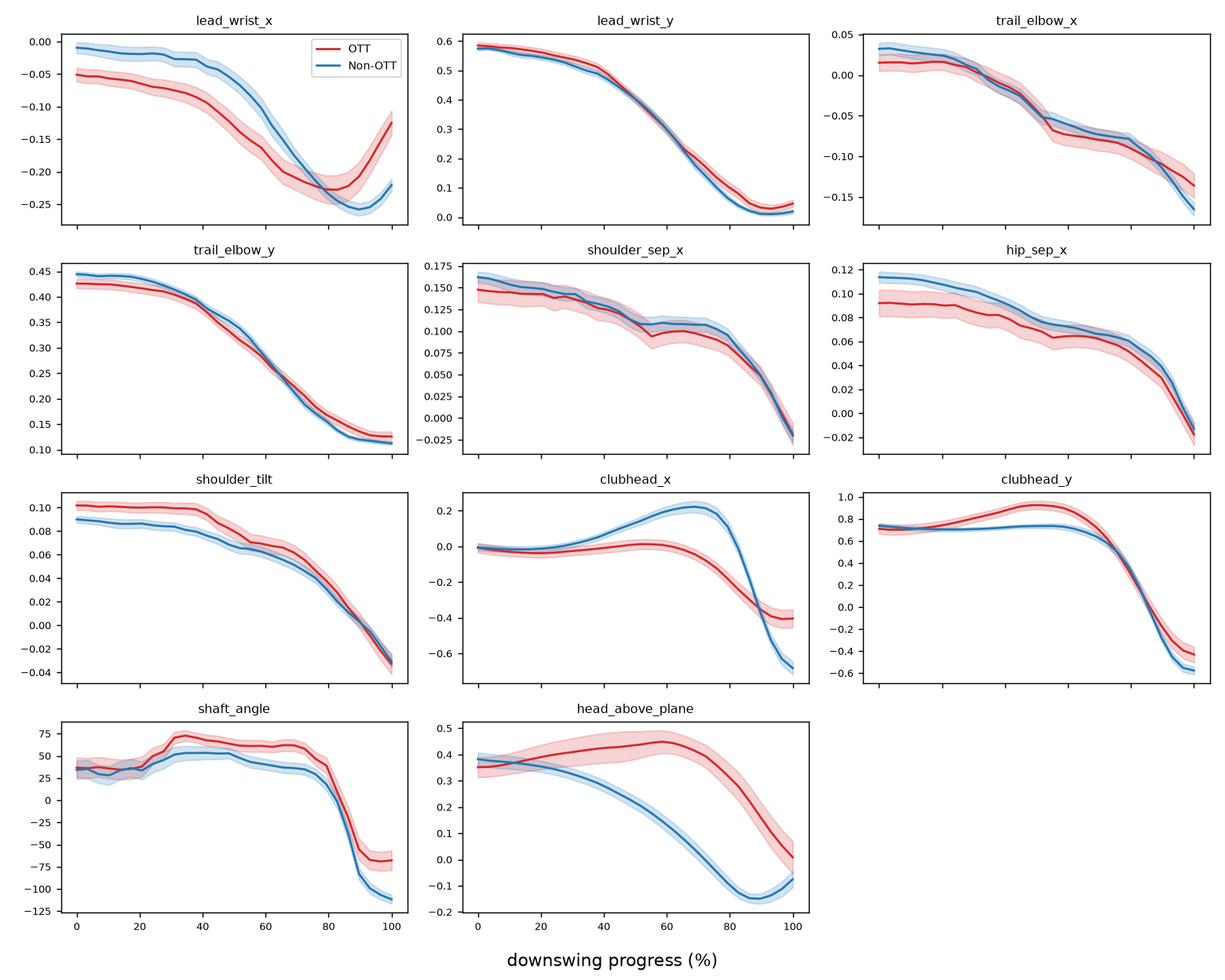}
\caption{Class-mean trajectories of the eleven OTT signals over the downswing (0\% is the top of the backswing, 100\% impact), OTT (red) against non-OTT (blue), with shaded bands around each mean showing the standard deviation. Positions are body-frame coordinates in body heights (Sec.~\ref{sec:norm}), the shaft angle is in degrees, and ``head\_above\_plane'' is the signed height of the clubhead above the swing-plane line (Sec.~\ref{sec:plane}).}
\label{fig:curves}
\end{figure}

\newpage
\section{Comparison of our model to a learned video model}
\label{app:video}
Another alternative to tracking is a learned video model. We tested a Kinetics-pretrained R3D-18 as a frozen feature extractor on each downswing. It reaches AUC 0.967, close to the XGBoost model. To test what each model relies on, we freeze every test swing at the top of the backswing, before the fault occurs, and re-score with each model. The R3D-18 still scores 0.946, separating the classes from appearance. Our model falls to 0.547, close to chance. Frame differences of the RGB video do not remove the shortcut either (0.951). On an earlier version of the data, Grad-CAM on the same backbone fed rendered skeleton frames placed its attention on the torso rather than the club or arms.

\begin{table}[H]
\centering
\small
\begin{tabular}{lccc}
\toprule
Model & Normal & \multicolumn{2}{c}{Frozen at top of backswing} \\
 & & re-scored & single-frame model \\
\midrule
R3D-18 (frozen) + log.\ reg., RGB & 0.967\,$\pm$\,0.028 & 0.946\,$\pm$\,0.040 & 0.960\,$\pm$\,0.034 \\
Ours (XGBoost, pos+delta)         & 0.975\,$\pm$\,0.019 & 0.547\,$\pm$\,0.077 & 0.752\,$\pm$\,0.055 \\
\bottomrule
\end{tabular}
\caption{Static-frame probe (CV ROC-AUC, 160 swings, 15 folds).
``Re-scored'': trained on real swings, tested on swings frozen at the top
of the backswing. ``Single-frame model'': trained and tested on that frame
alone.}
\label{tab:video}
\end{table}

\end{document}